\documentclass[preprint,12pt]{elsarticle}
\pdfoutput=1 

\usepackage{amssymb}
\usepackage{amsmath}
\usepackage{todonotes}
\usepackage{rotating}
\usepackage{subcaption}
\usepackage{algorithm}
\usepackage{algpseudocode}

\journal{Applied Mathematics and Computation}

\begin{document}

\begin{frontmatter}



\title{Image Trinarization Using a Partial Differential Equations: A Novel Approach to Automatic Sperm Image Analysis}


\author[1,2]{B. A. Jacobs \corref{cor1}}
\ead{byronj@uj.ac.za}
\cortext[cor1]{Corresponding author}
\address[1]{University of Johannesburg, Department of Mathematics and Applied Mathematics, South Africa}
\address[2]{Vitruvian Medical Diagnostics, YTC Building 02-01, 33 Maude Road, 208344, Singapore}

\begin{abstract}
Partial differential equations have recently garnered substantial attention as an image processing framework due to their extensibility, the ability to rigorously engineer and analyse the governing dynamics as well as the ease of implementation using numerical methods. This paper explores a novel approach to image trinarization with a concrete real-world application of classifying regions of sperm images used in the automatic analysis of sperm morphology. The proposed methodology engineers a diffusion equation with non-linear source term, exhibiting three steady-states. The model is implemented as an image processor using a standard finite difference method to illustrate the efficacy of the proposed approach. The performance of the proposed approach is benchmarked against standard image clustering/segmentation methods and shown to be highly effective.
\end{abstract}




\end{frontmatter}


\section{Introduction}\label{sec:Intro}
\begin{figure}[h!]
		\centering
		\begin{subfigure}{0.39\textwidth}
			\includegraphics[width=\textwidth]{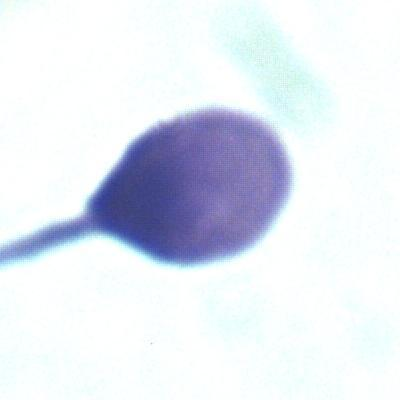}
			\caption{Example of Sperm Microscopy Image.}
			\label{fig:spermExample}
		\end{subfigure}
		\hfill
		\begin{subfigure}{0.6\textwidth}
			\includegraphics[width=0.95\linewidth]{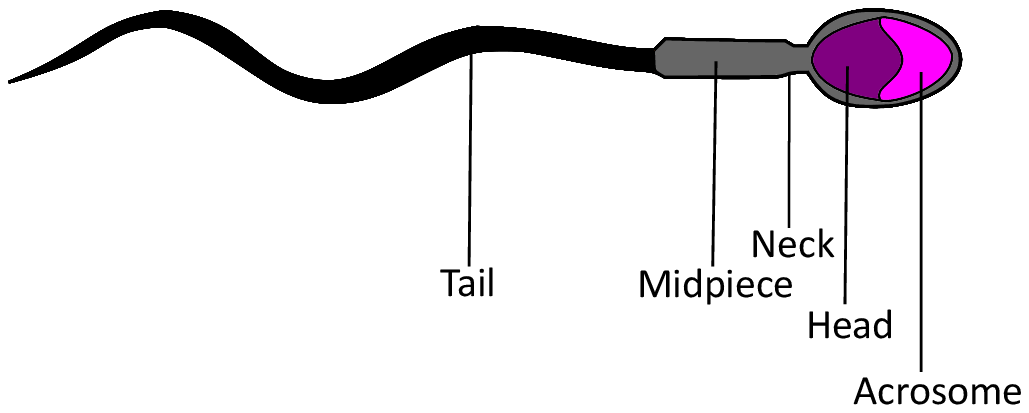}
			\caption{Diagram illustrating sperm anatomy.}
			\label{fig:spermAnatomy}
		\end{subfigure}
		\hfill
	\end{figure}

	Partial differential equations (PDEs) are a well established mathematical modelling tool and have been proven to show enormous value in a multitude of disciplines. More recently, PDEs have become a useful tool in image processing. Applications in image denoising, restoration, inpainting, remote sensing applications, document image binarization, \cite{perona1990scale, liu2011remote, kim2006pde, catte1992image, jacobs2013novel,jacobs2015locally, jacobs2018application,jacobs2022unsupervised, weeratunga2003comparison, chan2005image}, are a small subset of applications that illustrate the value of using a PDE framework to attain novel image processing algorithms. Additionally, research on processing medical images, such as ultrasounds and MRIs \cite{hadri2021novel, xing2011pde, lu2009four}.\\
	\\
	This work builds on previous works by the author \cite{jacobs2013novel,jacobs2015locally,jacobs2022unsupervised} where PDE models were used for document image binarization, implementing global thresholding, locally adaptive thresholding and a nonlinear system of equations which self-governed the model parameters resulting in an unsupervised methodology. The approach was also extended to time fractional PDEs presented in \cite{jacobs2018application}. Many other researchers have also contributed to the field using a variety of PDE based approaches. Feng \cite{feng2019novel, feng2022effective} presents variational level set approaches to the problem. An indirect diffusion approach to image segmentation is presented in \cite{wang2019indirect}, while Guo et al. \cite{guo2020fourth} show how indirect diffusion with shock filters can be used for text binarization. Zhang et al. \cite{zhang2020selective} present a nonlinear reaction diffusion model which exhibits the Perona-Malik \cite{perona1990scale} type diffusion (an edge preserving diffusion) coupled with a binarizing reaction term. An interger/fractional approach is combined with an edge detection to binarize heavily degraded document images \cite{nnolim2021dynamic, nnolim2021improved}. Du and He \cite{du2021nonlinear} propose a nonlinear diffusion model with selective source term which selectively restores the background or preserves the fidelity of the text.\\
	\\
	The task of image or data trinarization has received significantly less attention. The task aims to segment image data into three distinct classes corresponding to some desirable segmentation. This problem can be viewed as a base case of image segmentation, or component clustering and hence is amenable to standard approaches. Specifically K-Means \cite{macqueen1967some}, K-Medoids \cite{kaufman2009finding}, Spanning Trees \cite{asano1988clustering}, Agglomerative clustering \cite{rokach2005clustering}. This work benchmarks the proposed PDE approach against these standard approaches and illustrates that the fit-for-purpose PDE approach vastly outperforms the generalist approaches.\\
	\\
	Kido and Higo \cite{kido2020quantification} present the application of water retention tests, the results of which are amenable to data trinarization. The various phases of the test matter are trinarized using their approach presented in \cite{kido2017evaluation}. This technique is based on the frequency selection based on the image histogram data, a natural extension of Otsu's \cite{otsu1975threshold} technique for image binarization. Another approach taken by Higo et al. \cite{higo2014trinarization} for this application is the implementation of a region growing method. M\"ussel et al. \cite{mussel2016bitrina} present a multiscale approach for the binarization and trinarization of gene expression profiles. This application is motivated by the requirement of classifying data as `high', `low' or potentially `intermediate', which is particularly pertinent for biological data.\\ 
	\\
	Manual assessment of sperm morphology is a subjective process, which is prone to inter-expert variance and assessments are difficult to reproduce \cite{miahi2022genetic}. To address this problem computer aided semen analysis (CASA) has been an active research field for decades and has aimed to mitigate the subjectivity of the manual process. A natural approach to tackling the classification task of sperm morphology from images is by way of neural networks. Miahi et al. \cite{miahi2022genetic} present a novel Genetic Neural Architecture for the automatic assessment of sperm images. Javadi and Mirroshandel \cite{javadi2019novel} provide a deep learning approach, applying a convolutional deep neural network to the Sperm Morphology Analysis dataset (MHSMA). Savkay et al. \cite{csavkay2014sperm} present a convolutional neural network approach to CASA, presenting various morphological parameter predictions. An improved convolutional deep neural network approach is presented in \cite{prabaharan2021improved}. A comprehensive review of artificial intelligence in fertility is presented by Riegler et al. \cite{riegler2021artificial} which covers the broader subject matter of assisted reproductive technology, but collates key contributions to CASA using artificial intelligence. Dai et al. \cite{dai2021advances} present a review of recent advances in sperm analysis, illustrating that the field is an active one, garnering the attention of many researchers.\\
	\\
	The recent advances in CASA is largely focused on machine learning approaches. These techniques are well established as effective tools, but do suffer from some drawbacks typically associated with `black-box' methods. Additionally, the estimation of the acrosomal region is a key factor in the assessment of human sperm and is inherently subjective. According to the WHO Laboratory Manual for the Examination and Processing of Human Semen \cite{world2010laboratory} guidelines for sperm analysis this region should comprise 40\% to 70\% of the total sperm head. Figures \ref{fig:spermExample} and \ref{fig:spermAnatomy} illustrate an example of the sperm head and acrosomal region as well as a schematic of the sperm anatomy.\\
	\\	
	\\
	Building on the PDE based framework for image processing, the task of image trinarization is approached with the same lens. This implementation is motivated by the application of automatic sperm analysis from mircosopic images.	Establishing this PDE based framework as an image processing tool gives credence to the method for tackling new data driven problems. In the case of morphological sperm analysis, the dynamic nature of the PDE clustering method allows for each region to dynamically transit to its final classification. As has been shown before, this results in a method which is robust to noise, extensible and accurate.\\
	\\
	The following section introduces the proposed model and discusses the methodologies used to generate the results for automated sperm segmentation which are then discussed in Section \ref{sec:Results}. The paper concludes in Section \ref{sec:Conclusion} with insights given by the results and also how the present work opens new possibilities within the field of image processing with PDEs.
\section{Methods}\label{sec:Methods}
	\subsection{Model}
 
\begin{figure}[h!]
			\centering
			\includegraphics[width=0.95\linewidth]{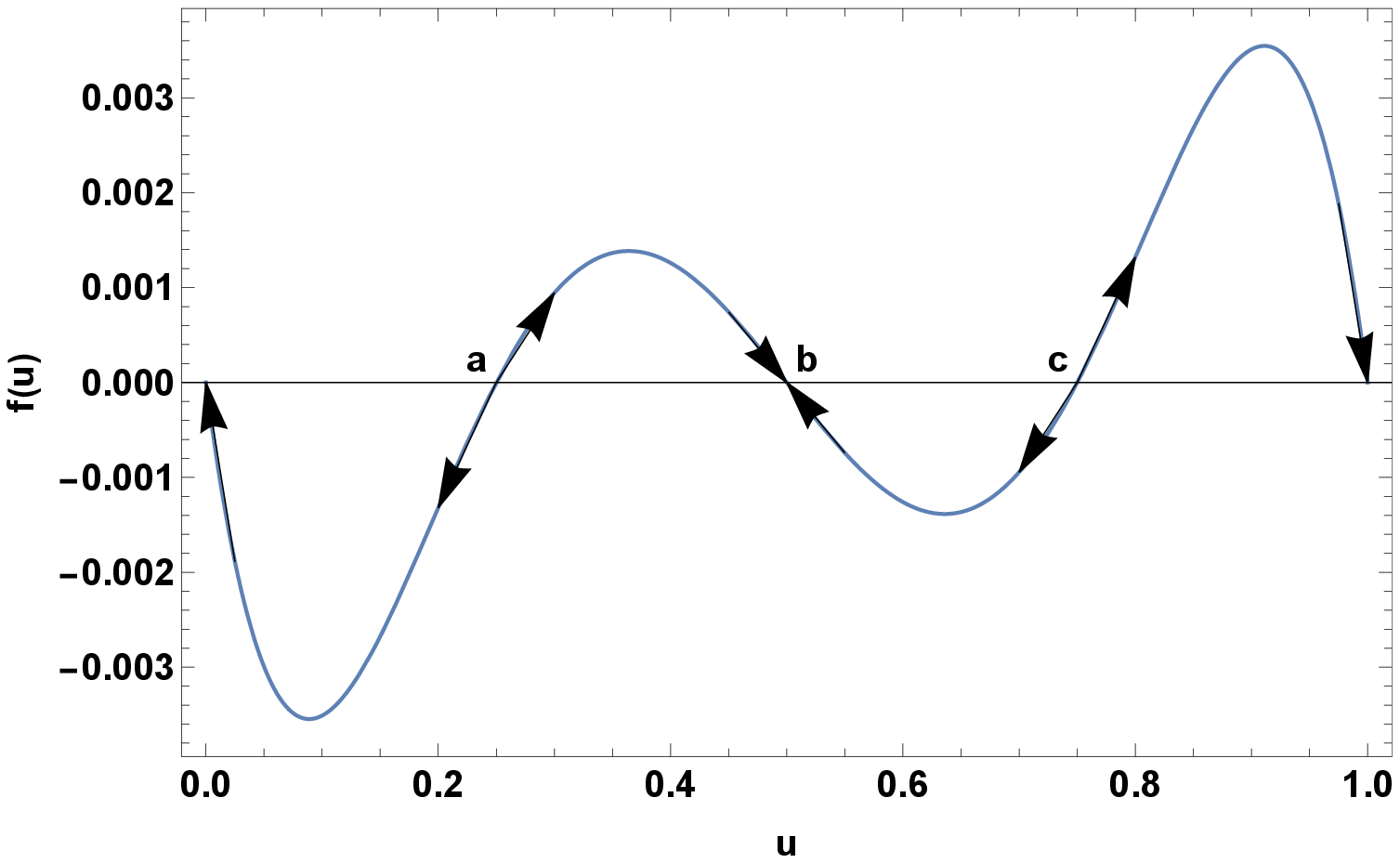}
			\caption{Phase space plot of the nonlinear source term, for $a=0.25$, $b=0.5$ and $c=0.75$.}
			\label{fig:sourceTerm}
		\end{figure}

	
		The model, given below, is engineered to exhibit three steady states, corresponding to background, sperm head and acrosomal region. Explicitly, the model is defined by
		\begin{align}
			u_t &= c_D \nabla u + c_S u(u-1)(u-a)(u-b), \label{eq:model} \\
			u(x,y,0) &= \text{Image}(x,y). \nonumber
		\end{align}
		The given model is constructed in two spatial dimensions, requiring that the input image be grayscale. Although, the extension to RGB images is straightforward under the PDE framework. The model is also subjected to zero-flux boundary conditions along all boundaries of the domain. This choice of boundary condition ensures smoothness at the boundaries and is motivated by previous works \cite{jacobs2013novel,jacobs2015locally,jacobs2022unsupervised}. Explicitly, the boundary conditions are given by,\\
		\begin{align}
			u_x(0,y,t) = 0 \nonumber\\
			u_x(1,y,t) = 0 \nonumber\\
			u_y(x,0,t) = 0 \nonumber\\
			u_y(x,1,t) = 0 \nonumber
		\end{align}
		\\
		In engineering the dynamics of the present model, we require the model to exhibit three stable steady states. This is achieved by the careful construction of the source term. Figure \ref{fig:sourceTerm} illustrate the phase space dynamics of the source term. This structure is an extension of the Fitzhugh-Nagumo model used in previous works \cite{jacobs2013novel,jacobs2015locally,jacobs2022unsupervised}. The source term exhibits five steady states; $u=0, u=a, u=b, u=c$ and $u=1$. Assuming that $0 \le a < b < c\le 1$, the stable states exist at $u=0, u=1$ and $u=b$ as illustrated in Figure \ref{fig:sourceTerm}.
		
		Moreover, due to the nature of the input images, specifically high magnification with uniform illumination, there is likely minimal noise captured in the image which is associated with the camera sensor. The coefficients $c_D$ and $c_S$ are tunable to balance the bias of smoothing (denoising) versus trinarization.\\
		\\
		
	\subsection{Numerical Method}
		The standard finite-difference implementation offers a conditionally stable, explicit and parallelizable \cite{jacobs2015locally} method. Although the method is rudimentary its ease of implementation offers an attractive option for robust experimentation. The model is discretised as follows:
		\begin{align}
			u(x_i,y_j,t_n) &= u(i \Delta x, j \Delta y, n \Delta n) = u_{i,j}^n, \\
			u_t(x,y,t) &= \frac{u_{i,j}^{n+1}-u_{i,j}^{n}}{\Delta t} + \mathcal{O}(\Delta t),  \\
			u_x(x,y,t) &= \frac{u_{i+1,j}^{n}-u_{i-1,j}^{n}}{2\Delta x} + \mathcal{O}(\Delta x^2), \label{eq:BCx}  \\			
			u_y(x,y,t) &= \frac{u_{i,j+1}^{n}-u_{i,j-1}^{n}}{2\Delta y} + \mathcal{O}(\Delta y^2), \label{eq:BCy}\\	
			u_{xx}(x,y,t) &= \frac{u_{i+1,j}^{n}-2u_{i,j}^{n}+u_{i-1,j}^{n}}{\Delta x^2} + \mathcal{O}(\Delta x^2),  \\
			u_{yy}(x,y,t) &= \frac{u_{i,j+1}^{n}-2u_{i,j}^{n}+u_{i,j-1}^{n}}{\Delta y^2} + \mathcal{O}(\Delta y^2). 
		\end{align}
		The computational domain is prescribed by the input image, but is artificially normalised to $(x,y) \in [0,1]\times [0,1]$. The image dimensions, $N$ and $M$, thus dictate $\Delta x = 1/(N-1)$ and $\Delta y = 1/(M-1)$. Discretising the model in \eqref{eq:model} yields the explicit update scheme,
		\begin{align}
			u_{i,j}^{n+1} = u_{i,j}^{n} &+ \frac{\Delta t c_D}{\Delta x^2} \left(u_{i+1,j}^{n}-2u_{i,j}^{n}+u_{i-1,j}^{n}\right) \nonumber \\
			 &+ \frac{\Delta t c_D}{\Delta y^2} \left(u_{i,j+1}^{n}-2u_{i,j}^{n}+u_{i,j-1}^{n}\right) \nonumber \\
			 &- \Delta t c_S u_{i,j}^n (u_{i,j}^n-1)(u_{i,j}^n-a)(u_{i,j}^n-b)(u_{i,j}^n-c). \label{eq:numerics}
		\end{align}
		The resulting scheme is consistent with the model equation \eqref{eq:model} as $\Delta x, \Delta y, \Delta t \rightarrow 0$. \\
		\\
		The boundary conditions are discretised using the central difference formulae \eqref{eq:BCx} and \eqref{eq:BCy} to ensure second order accuracy at the boundaries. Explicitly we have
		\begin{align}
			u_x(0,y,t) &= 0 \Rightarrow u_{-1,j}^{n}=u_{1,j}^{n}, \\
			u_x(1,y,t) &= 0 \Rightarrow u_{N+1,j}^{n}=u_{N-1,j}^{n}, \\
			u_y(x,0,t) &= 0 \Rightarrow u_{j,-1}^{n}=u_{j,1}^{n}, \\
			u_y(x,1,t) &= 0 \Rightarrow u_{M+1,j}^{n}=u_{M-1,j}^{n}. 
		\end{align}
		For simplicity, the scheme can be written in matrix form
		\begin{align}
			{\bf u}^{n+1} &= {\bf u}^{n+1} + \frac{cD \Delta t}{\Delta x^2} {\bf A}.{\bf u}^{n} + \frac{c_D \Delta t}{\Delta y^2} {\bf u}^{n}.{\bf A}^T \nonumber \\
			&- c_S  \Delta t {\bf u}^n \circ ({\bf u}^n - 1)\circ  ({\bf u}^n - a) \circ ({\bf u}^n - b)\circ  ({\bf u}^n - c),
		\end{align}
		where $\circ$ denotes the Hadamard product and
		\begin{equation}
			{\bf u} = \left(
			\begin{array}{ccccc}
				u_{0,0}^n & u_{0,1}^n & \ldots & u_{0,M-1}^n & u_{0,M}^n \\
				u_{1,0}^n & u_{1,1}^n &   & u_{1,M-1}^n &  u_{1,M}^n\\
				\vdots &  &\ddots&  & \vdots \\
				u_{N-1,0}^n & u_{N-1,1}^n &   & u_{N-1,M-1}^n & u_{N-1,M}^n \\
				u_{N,0}^n & u_{N,1}^n &\ldots& u_{N,M-1}^n & u_{N,M}^n \\
			\end{array}
			\right),
		\end{equation}
		\begin{equation}
			{\bf A} = \left(
			\begin{array}{ccccccc}
				-2 & 2 & 0 & \ldots & 0 & 0 & 0 \\
				1 & -2 & 1 &  & 0 & 0 & 0 \\
				0 & 1 & -2 &  & 0 & 0 & 0 \\
				\vdots &  &  & \ddots &  &  & \vdots \\
				0 & 0 & 0 &   & -2 & 1 & 0 \\
				0 & 0 & 0 &   & 1 & -2 & 1 \\
				0 & 0 & 0 & \ldots & 0 & 2 & -2 \\
			\end{array}
			\right).
		\end{equation}

	\subsection{Stability Analysis}
		To accommodate the nonlinearity in the stability analysis, the source term is replaced by the worst case contribution. This leads to stability conditions which are stricter than necessary in most cases. Nevertheless, the stability requirement does not overly impose on the functioning the method.\\
		Consider the source term in \eqref{eq:model},
		\begin{equation}
			f(u) = -u(u-1)(u-a)(u-b)(u-c).
		\end{equation}
		This function is maximal (in absolute value) under constraints $0 \le u \le 1$, $0 \le a \le 1$ and $0 \le b \le 1$ with 
		\begin{equation}
			\max_{u,a,b,c} |f(u)| = \frac{256}{3125}.
		\end{equation}
		The von Neumann stability analysis is conducted by introducing the ansatz 
		\begin{equation}
			u_{i,j}^n = \zeta^n e^{I i\Delta x j \Delta y \omega},
		\end{equation}
		where $\omega$ indicates an arbitrary Fourier wave number and $I = \sqrt{-1}$. Deriving the conditions on $\Delta x$, $\Delta y$ and $\Delta t$ that ensure the amplification factor, $\zeta$, is smaller than one in absolute value, $|\zeta|< 1 $, is sufficient to guarantee the numerical method is stable. Substituting the ansatz into equation \eqref{eq:numerics} we have
		\begin{align}
			\zeta = 1 &-\frac{2\Delta t c_D}{\Delta x^2} + \frac{2\Delta t c_D}{\Delta x^2} \frac{e^{I \Delta x j \Delta y \omega} + e^{-I\Delta x j \Delta y \omega}}{2} \nonumber \\
			&-\frac{2\Delta t c_D}{\Delta y^2} + \frac{2\Delta t c_D}{\Delta y^2} \frac{e^{I i\Delta x \Delta y \omega} + e^{-I i\Delta x \Delta y \omega}}{2} + \Delta t c_S f(u_{i,j}^n), \nonumber \\
			= 1 &+\frac{2\Delta t c_D}{\Delta x^2} \left(\cos\left(2\frac{ j \Delta x \Delta y \omega}{2}\right)-1\right) \nonumber \\
			&+\frac{2\Delta t c_D}{\Delta y^2} \left(\cos\left(2\frac{ i \Delta x \Delta y \omega}{2}\right)-1\right)+ \Delta t c_S f(u_{i,j}^n), \nonumber \\
			= 1 &-\frac{2\Delta t c_D}{\Delta x^2} \sin^2\left(j \Delta x \Delta y \omega\right) \nonumber \\
			&-\frac{2\Delta t c_D}{\Delta y^2} \sin^2\left(i \Delta x \Delta y \omega\right) + \Delta t c_S f(u_{i,j}^n). 
		\end{align}
		The stability condition $|\zeta| <1$ leads to 
		\begin{align}
			\Delta t c_S |f(u_{i,j}^n)| &\le \Delta t c_S \frac{256}{3125}, \nonumber\\
			&<\frac{2\Delta t c_D}{\Delta x^2} \sin^2\left(j \Delta x \Delta y \omega\right) +\frac{2\Delta t c_D}{\Delta y^2} \sin^2\left(i \Delta x \Delta y \omega\right), \nonumber \\
			&<2\Delta t c_D \left(\frac{1}{\Delta x^2} +\frac{1}{\Delta y^2} \right), \nonumber \\
		\end{align}
		and
		\begin{align}
			-2 &< -\frac{2\Delta t c_D}{\Delta x^2} \sin^2\left(j \Delta x \Delta y \omega\right) -\frac{2\Delta t c_D}{\Delta y^2} \sin^2\left(i \Delta x \Delta y \omega\right) + \Delta t c_S f(u_{i,j}^n) \nonumber \\
			&\le -\frac{2\Delta t c_D}{\Delta x^2} - \frac{2\Delta t c_D}{\Delta y^2} + \Delta t c_S f(u_{i,j}^n) \nonumber \\
		2\Delta t c_D \left(\frac{1}{\Delta x^2} + \frac{1}{\Delta y^2} - \frac{1}{\Delta t c_D }\right) & \le \Delta t c_S |f(u_{i,j}^n)| \le \Delta t c_S \frac{256}{3125}. 
		\end{align}
		The above criteria can be summarised as
		\begin{equation}\label{eq:stabilityCond}
			|\Delta t c_S \frac{128}{3125}| \le c_D \left(\frac{\Delta t }{\Delta x^2} +\frac{\Delta t }{\Delta y^2} \right).
		\end{equation}
		Since $\Delta x$ and $\Delta y$ are prescribed by the input image's dimensions, $\Delta t$ should be chosen so that $\frac{\Delta t}{\Delta x^2}$ and $\frac{\Delta t}{\Delta y^2}$ are $\mathcal{O}(1)$. Then $c_S$ and $c_D$ can be chosen to balance the effect of trinarization and smoothing respectively, while still obeying equation \eqref{eq:stabilityCond}, thus ensuring the resulting implementation is stable.
\section{Medical Image Application}\label{sec:Results}
	This section presents the performance results of the present method as well as the benchmark figures for standard methodologies for image segmentation. Performance metrics are discussed which enable the quantitative measurement of performance of various methodologies. The application of segmentation of sperm head and acrosomal regions is an important aspect of CASA and has been an active field of research for many years. This work presents the novel approach of performing this segmentation using a PDE based framework. The results presented here illustrate that the PDE based approach allows for a PDE model to be engineered to exhibit desirable dynamics required by the application, and more importantly, provide a robust, provably stable and highly accurate method for CASA.\\
	\\
	The images of sperm obtained from microscopy shown in Figures \ref{fig:ex1in}, \ref{fig:ex2in}, \ref{fig:ex3in}, \ref{fig:ex4in} and \ref{fig:ex5in}, illustrate the subjective nature of discerning exactly where the acrosomal regions begin and end. As such, a robust algorithm which provides repeatable results is of significant importance in CASA.
	\subsection{Algorithm Description}
	In order to improve the classification results a post processing step is implemented, which is specific to the CASA application. Specifically this optimisation removes extraneous information at the edges of the image, or trimming of the sperm tail by constructing a mask from the image using traditional image processing techniques. This post processing was done for all clustering methods for a fair comparison of the segmentation task. The overall algorithm is described in Algorithm \ref{alg:final}.
	\begin{algorithm}
		\caption{Segmentation and Post-Processing Algorithm}\label{alg:final}
		\begin{algorithmic}[1]
				\State $img \gets \text{Input Image}$
				\State $result \gets clusteringMethod(img)$ \Comment{from selected clustering method}
				\State $result \gets round(2*result)/2$ \Comment{force all pixel values to 0, 0.5 or 1}
				\State $mask \gets closing(res, disk(20))$ \Comment{image mask formed with disk kernel}
				\State $result \gets applyMask(result,mask)$ \Comment{Mask resulting image} 
				
		\end{algorithmic}
	\end{algorithm}
	\subsection{Performance Metrics}
		The quantitative performance of the assessed clustering methodologies requires a robust performance measure. The performance of each algorithm is broken down into two binary classification tasks; measuring the number of pixels classified as head and measuring the number of pixels classified as acrosome. The classification results are shown in Figures \ref{fig:ex1}, \ref{fig:ex2}, \ref{fig:ex3}, \ref{fig:ex4} and \ref{fig:ex5}, where class 1 is shown in grey and class 2 is shown in black.\\
		\\
		A natural approach to measuring the accuracy of a binary classification task is to count the number of pixels classified as black which are also classified as black in the ground truth image, and scale that count by the total number of pixels in the image; in essence giving a percentage of ``correctly classified" pixels. These results are presented in Table \ref{tab:resultsAccuracy} for each image and each method. However, this is biased in the case where the number of target pixels is much less than the total number of pixels in the entire image. For example, if the binary ground truth image contained only one black pixel, then a pure white image would be considered almost perfectly accurate, but in reality the importance of capturing that one black pixel is the core task. The overestimation of a method's performance is evident in the high accuracies obtained by all methods in Table \ref{tab:resultsAccuracy}\\
		\\
		As such, many researchers adopt the F-measure, or $F_1$ score \cite{taha2015metrics} as a more meaningful measure of accuracy in binary classification tasks. This measure is the harmonic mean of the precision and recall of the classification result, essentially compensating for the bias in class representation. The $F_1$ score is calculated as
		\begin{equation}
			F_1 = 2\frac{ \text{precision} \times \text{recall}  }{\text{precision} + \text{recall}} = \frac{\text{tp}}{\text{tp}+\frac{1}{2}\left(\text{fp}+\text{fn}\right)}.
		\end{equation}
		The $F_1$ scores obtained by all methods across all images are presented in Table \ref{tab:resultsF1Score} and it can be seen that these scores correlate more meaningfully with the qualitative results shown in Figures \ref{fig:ex1}, \ref{fig:ex2}, \ref{fig:ex3}, \ref{fig:ex4} and \ref{fig:ex5}.
		
	\subsection{Results}
\begin{figure}
	\centering
	\begin{subfigure}{0.49\textwidth}
		\includegraphics[width=\textwidth]{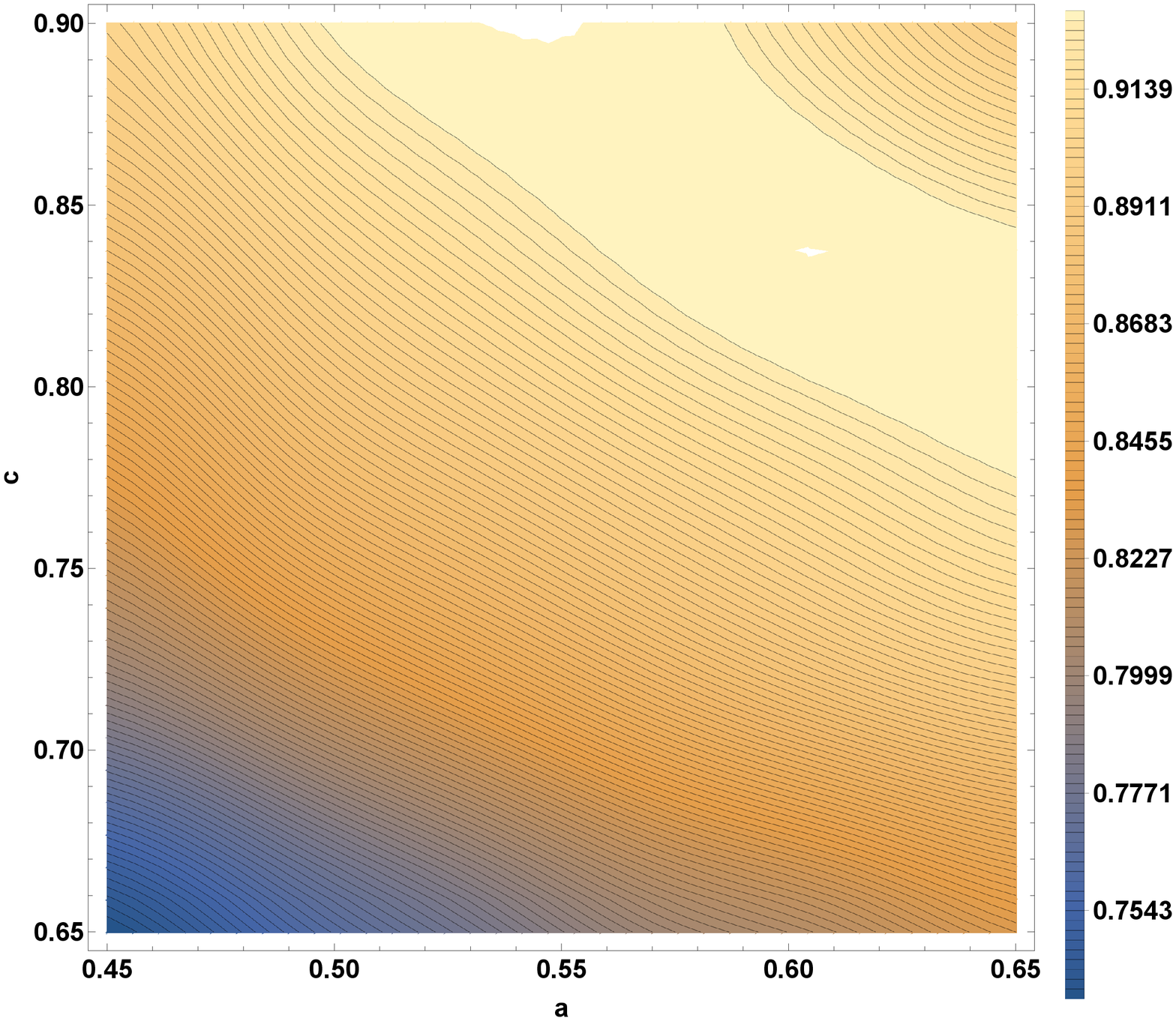}
		\caption{Contour error plot for varying $a$ and $c$ with fixed $b=0.65 $.}
		\label{fig:ex1Contour}
	\end{subfigure}
	\hfill
		\begin{subfigure}{0.49\textwidth}
		\includegraphics[width=\textwidth]{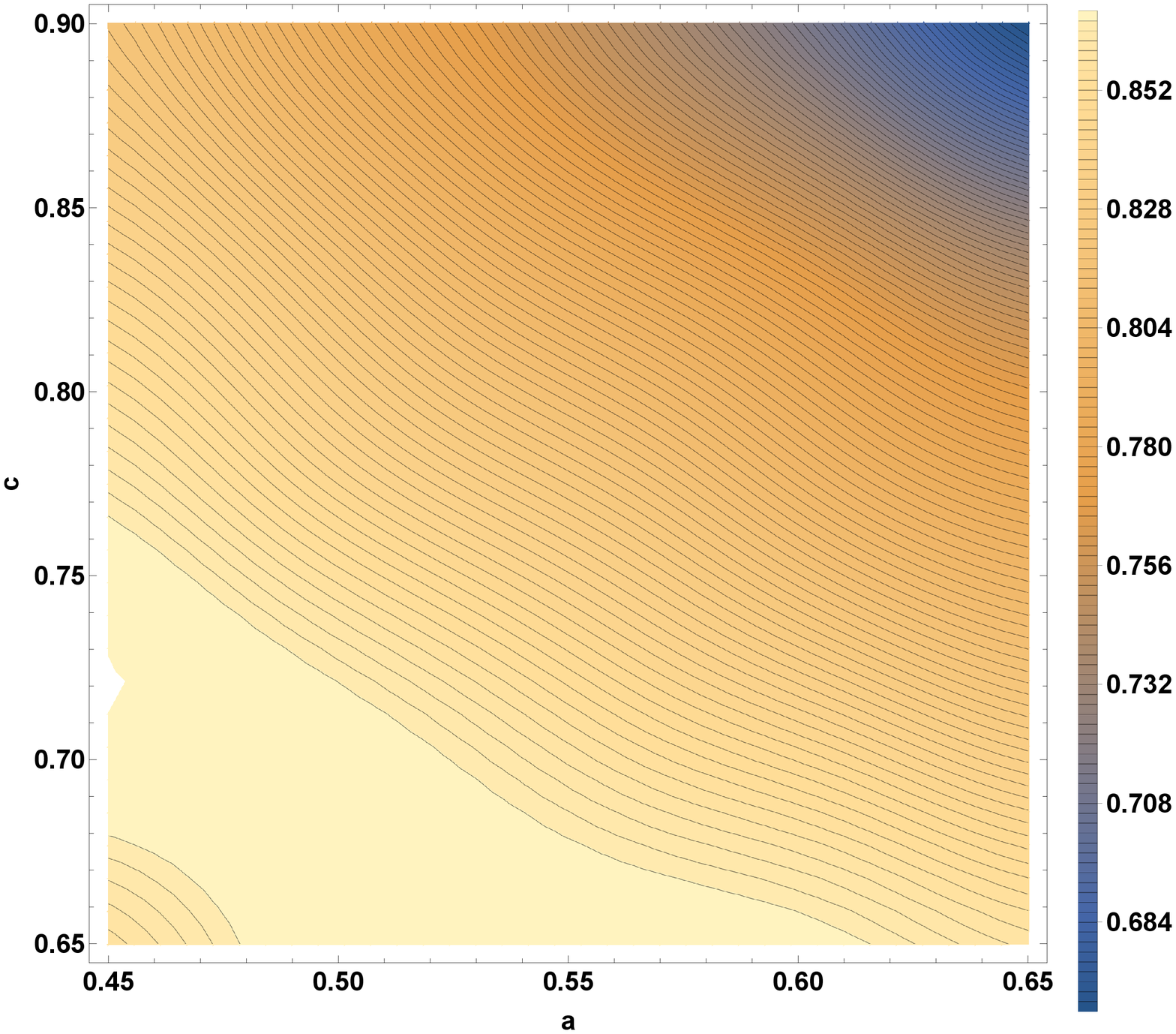}
		\caption{Contour error plot for varying $a$ and $c$ with fixed $b=0.65 $.}
		\label{fig:ex2Contour}
	\end{subfigure}
	\hfill
		\begin{subfigure}{0.49\textwidth}
		\includegraphics[width=\textwidth]{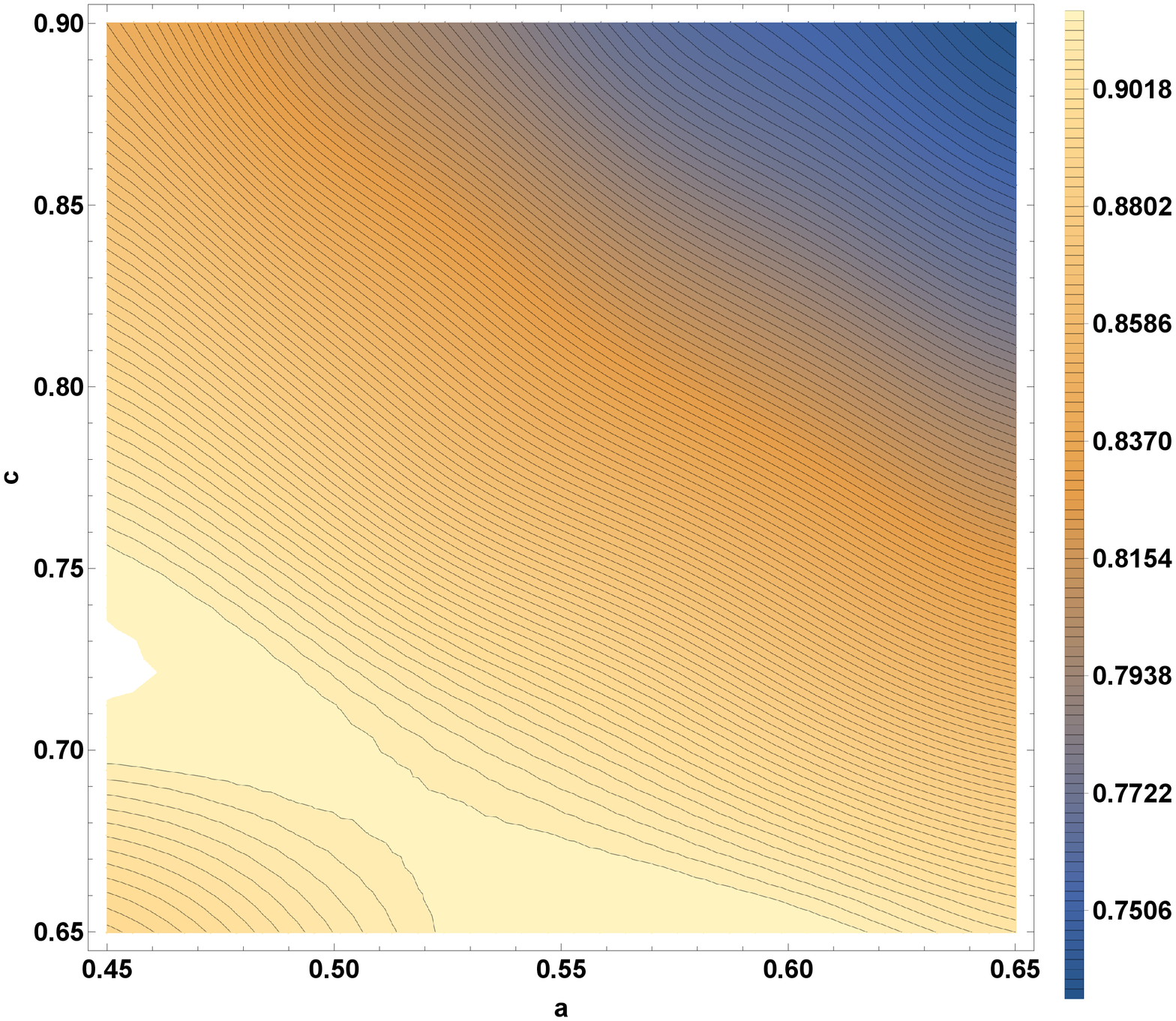}
		\caption{Contour error plot for varying $a$ and $c$ with fixed $b=0.65 $.}
		\label{fig:ex3Contour}
	\end{subfigure}
	\hfill
		\begin{subfigure}{0.49\textwidth}
		\includegraphics[width=\textwidth]{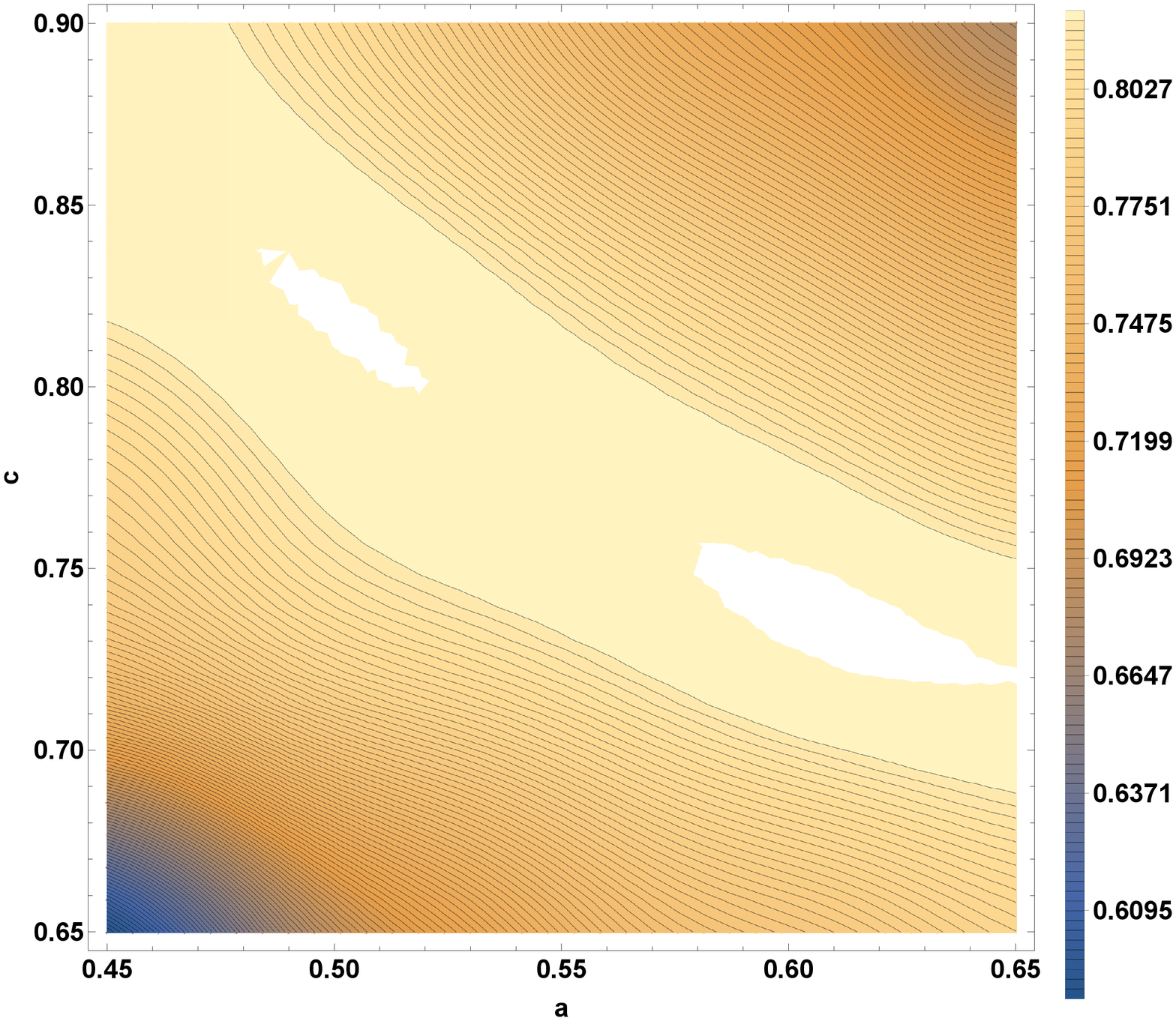}
		\caption{Contour error plot for varying $a$ and $c$ with fixed $b=0.65 $.}
		\label{fig:ex4Contour}
	\end{subfigure}
	\hfill
		\begin{subfigure}{0.49\textwidth}
		\includegraphics[width=\textwidth]{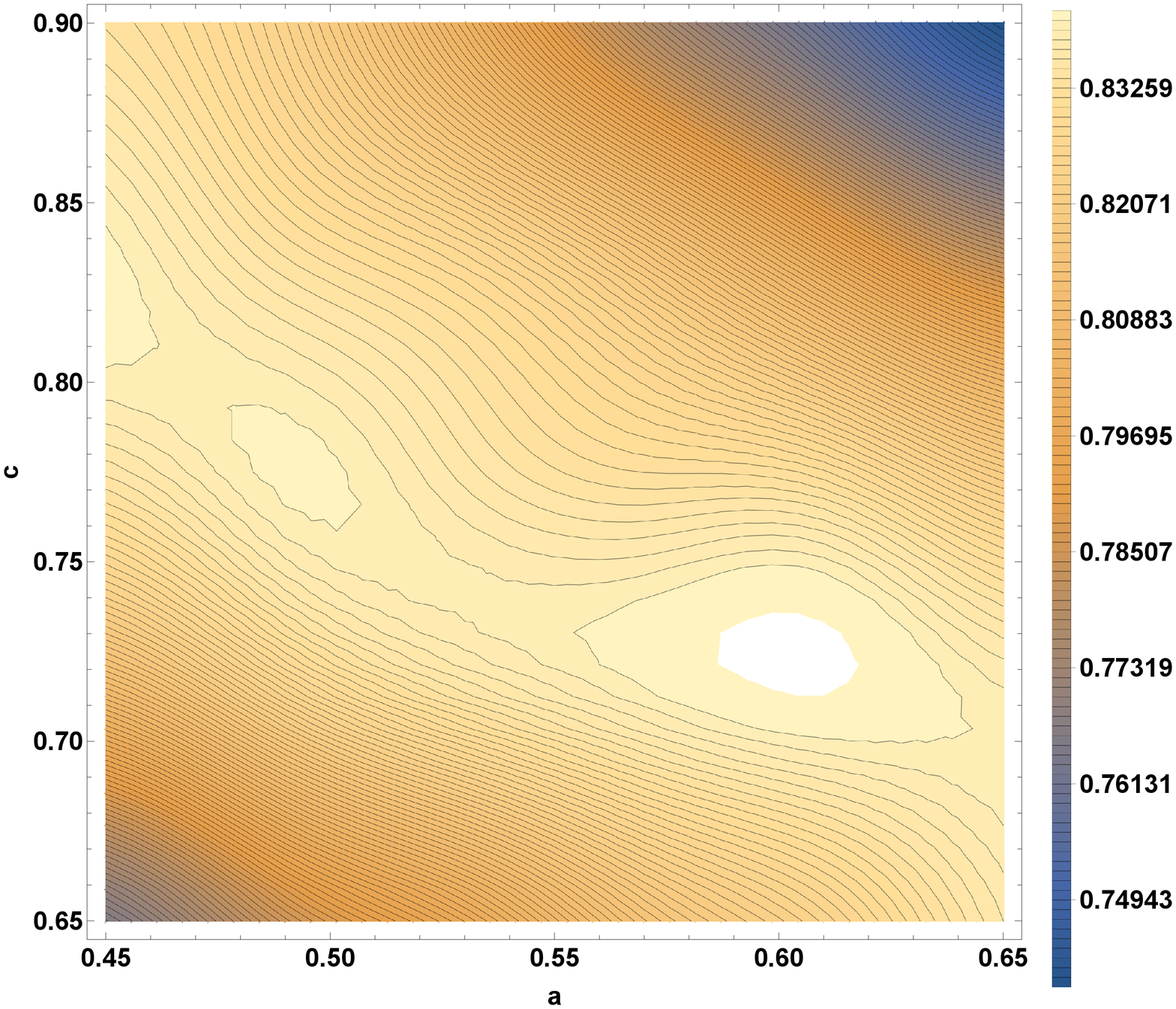}
		\caption{Contour error plot for varying $a$ and $c$ with fixed $b=0.65 $.}
		\label{fig:ex5Contour}
	\end{subfigure}
		\caption{Contour plots of $F_1$ scores for all example images for various values of $a$ and $c$.}
	\label{fig:contours}
	\end{figure}

	The proposed method relies on the tuning of parameters $a$, $b$ and $c$, however in the experiments presented below, the algorithm proved to be robust for all input images, with fixed parameters of $a = 0.5$, $b=0.65$ and $c=0.7$. As depicted in Figure \ref{fig:sourceTerm}, the value of $b$ determines the stable value to which pixels will be driven. Due to this the value of $b$ can be fixed and $a$ and $c$ varied to change the relative strength of the source term in driving pixel values toward $0$, $b$ or $1$.\\
	\\
	Typical values were selected for the remaining model parameters, namely $c_D = 0.01$, $c_S = 1/\Delta t$ and $\Delta t = \Delta x \Delta y/4$ where $\Delta x = 1/N$ and $\Delta y = 1/M$ with $(N,M)$ being the dimensions of the input image. The choice of these parameters is informed by previous studies, indicating that the relatively small value of $c_D$ is sufficient to introduce classification dynamics when the image is largely free of noise. The diffusion term encourages smoothing of noisy data and boundary regions.\\
	\\
	The $F_1$ scores as a function of $a$ and $c$ are plotted in Figure \ref{fig:contours}. These figures illustrate that near optimal results can be obtained from the present methodology using the aforementioned fixed parameter values.\\ 
	\\

\begin{figure}
		\centering
		\begin{subfigure}{0.32\textwidth}
			\includegraphics[width=\textwidth]{in_1.png}
			\caption{Input Image.}
			\label{fig:ex1in}
		\end{subfigure}
		\hfill
		\begin{subfigure}{0.32\textwidth}
			\includegraphics[width=\textwidth]{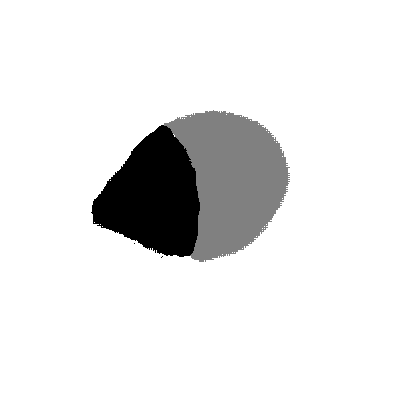}
			\caption{Ground Truth.}
			\label{fig:ex1GT}
		\end{subfigure}
		\hfill
		\begin{subfigure}{0.32\textwidth}
			\includegraphics[width=\textwidth]{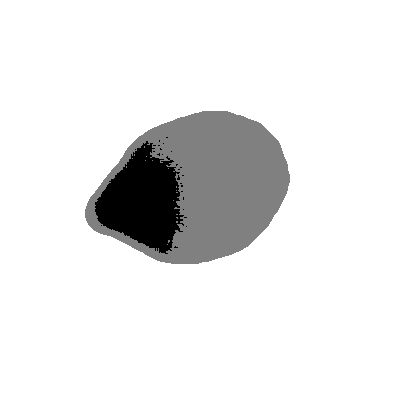}
			\caption{PDE Result.}
			\label{fig:ex1PDE}
		\end{subfigure}
		\hfill
		\begin{subfigure}{0.32\textwidth}
			\includegraphics[width=\textwidth]{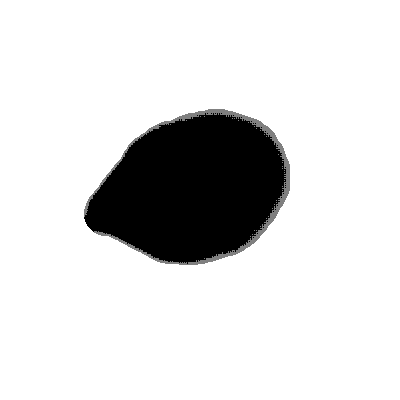}
			\caption{Agglomerative Clustering.}
			\label{fig:ex1Agg}
		\end{subfigure}
		\hfill
		\begin{subfigure}{0.32\textwidth}
			\includegraphics[width=\textwidth]{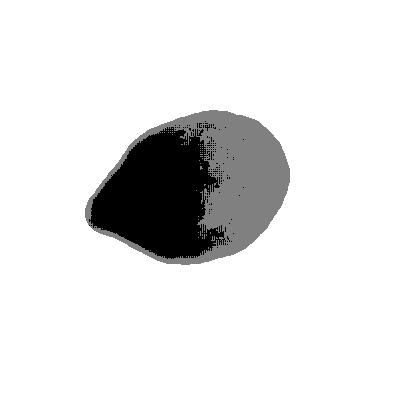}
			\caption{K Means Clustering.}
			\label{fig:ex1Kmean}
		\end{subfigure}
		\hfill
		\begin{subfigure}{0.32\textwidth}
			\includegraphics[width=\textwidth]{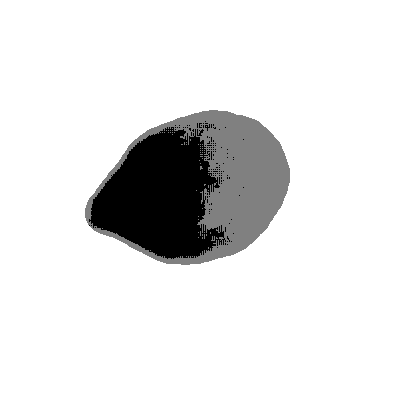}
			\caption{K Medoids Clustering.}
			\label{fig:ex1KMedoid}
		\end{subfigure}
		\hfill
		\begin{subfigure}{0.32\textwidth}
			\includegraphics[width=\textwidth]{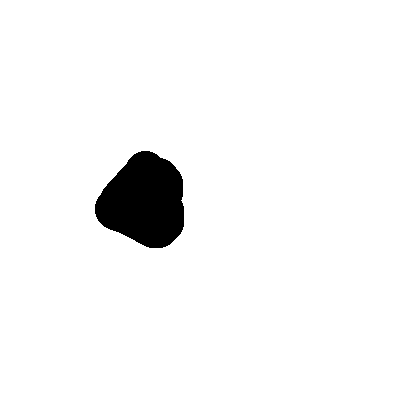}
			\caption{Spanning Trees.}
			\label{fig:ex1Spanning}
		\end{subfigure}
		\caption{Input image, ground truth and results from a variety algorithms for Example 1.}
		\label{fig:ex1}
	\end{figure}

\begin{figure}
		\centering
		\begin{subfigure}{0.32\textwidth}
			\includegraphics[width=\textwidth]{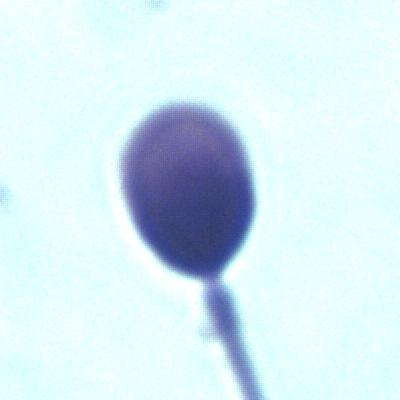}
			\caption{Input Image.}
			\label{fig:ex2in}
		\end{subfigure}
		\hfill
		\begin{subfigure}{0.32\textwidth}
			\includegraphics[width=\textwidth]{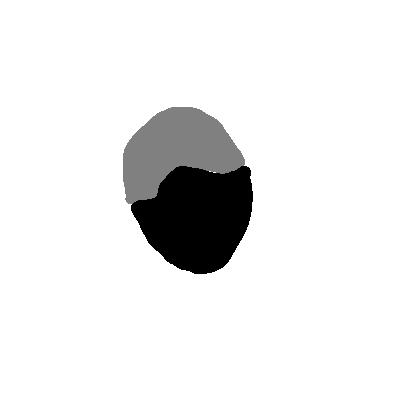}
			\caption{Ground Truth.}
			\label{fig:ex2GT}
		\end{subfigure}
		\hfill
		\begin{subfigure}{0.32\textwidth}
			\includegraphics[width=\textwidth]{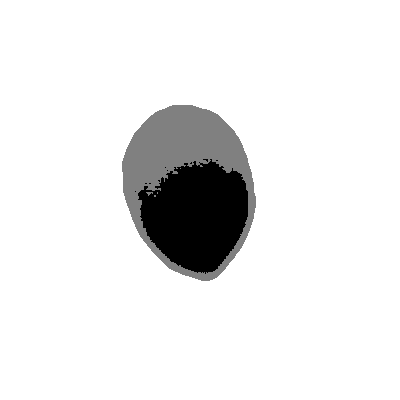}
			\caption{PDE Result.}
			\label{fig:ex2PDE}
		\end{subfigure}
		\hfill
		\begin{subfigure}{0.32\textwidth}
			\includegraphics[width=\textwidth]{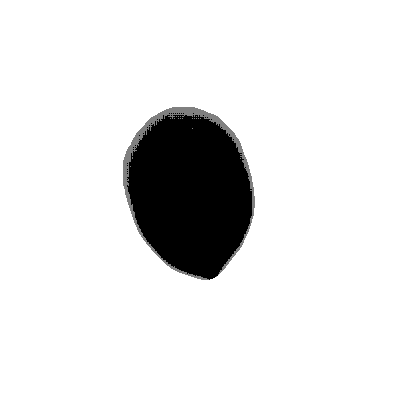}
			\caption{Agglomerative Clustering.}
			\label{fig:ex2Agg}
		\end{subfigure}
		\hfill
		\begin{subfigure}{0.32\textwidth}
			\includegraphics[width=\textwidth]{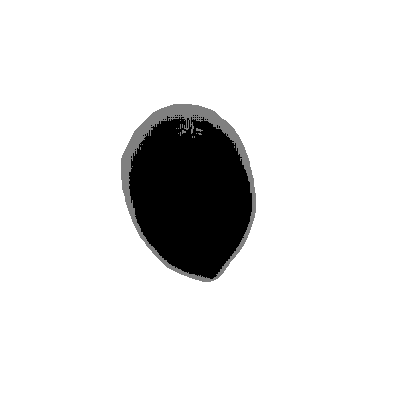}
			\caption{K Means Clustering.}
			\label{fig:ex2Kmean}
		\end{subfigure}
		\hfill
		\begin{subfigure}{0.32\textwidth}
			\includegraphics[width=\textwidth]{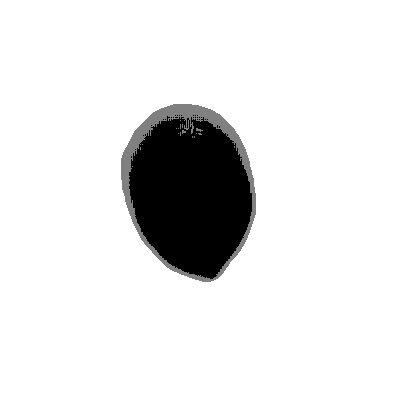}
			\caption{K Medoids Clustering.}
			\label{fig:ex2KMedoid}
		\end{subfigure}
		\hfill
		\begin{subfigure}{0.32\textwidth}
			\includegraphics[width=\textwidth]{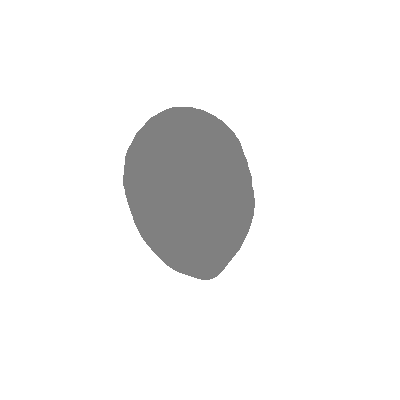}
			\caption{Spanning Trees.}
			\label{fig:ex2Spanning}
		\end{subfigure}
		\caption{Input image, ground truth and results from a variety algorithms for Example 2.}
		\label{fig:ex2}
	\end{figure}

\begin{figure}
		\centering
		\begin{subfigure}{0.32\textwidth}
			\includegraphics[width=\textwidth]{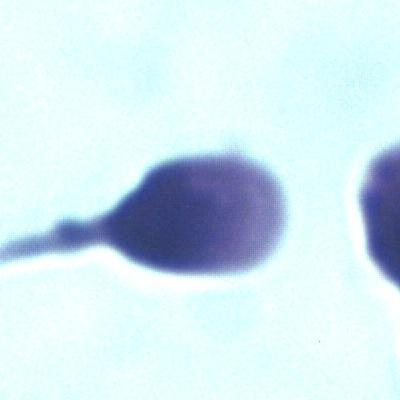}
			\caption{Input Image.}
			\label{fig:ex3in}
		\end{subfigure}
		\hfill
		\begin{subfigure}{0.32\textwidth}
			\includegraphics[width=\textwidth]{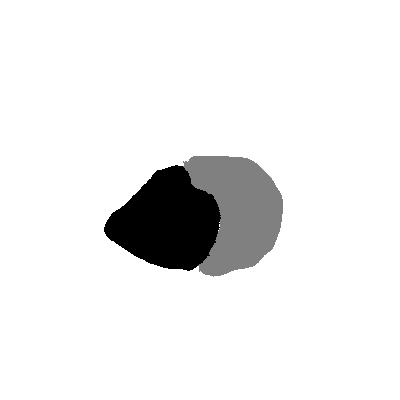}
			\caption{Ground Truth.}
			\label{fig:ex3GT}
		\end{subfigure}
		\hfill
		\begin{subfigure}{0.32\textwidth}
			\includegraphics[width=\textwidth]{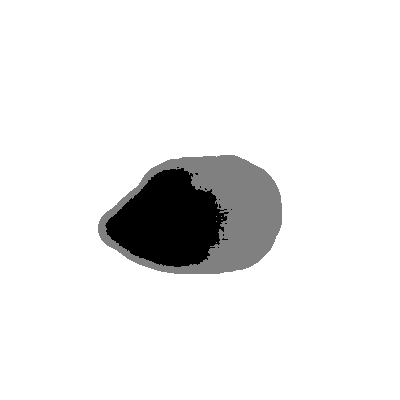}
			\caption{PDE Result.}
			\label{fig:ex3PDE}
		\end{subfigure}
		\hfill
		\begin{subfigure}{0.32\textwidth}
			\includegraphics[width=\textwidth]{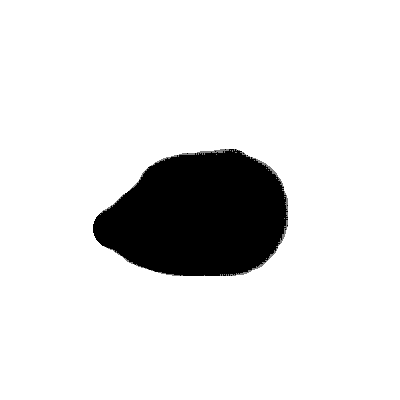}
			\caption{Agglomerative Clustering.}
			\label{fig:ex3Agg}
		\end{subfigure}
		\hfill
		\begin{subfigure}{0.32\textwidth}
			\includegraphics[width=\textwidth]{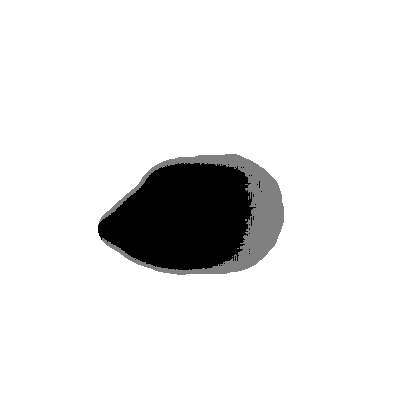}
			\caption{K Means Clustering.}
			\label{fig:ex3Kmean}
		\end{subfigure}
		\hfill
		\begin{subfigure}{0.32\textwidth}
			\includegraphics[width=\textwidth]{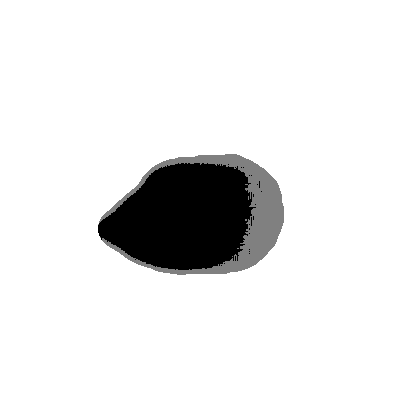}
			\caption{K Medoids Clustering.}
			\label{fig:ex3KMedoid}
		\end{subfigure}
		\hfill
		\begin{subfigure}{0.32\textwidth}
			\includegraphics[width=\textwidth]{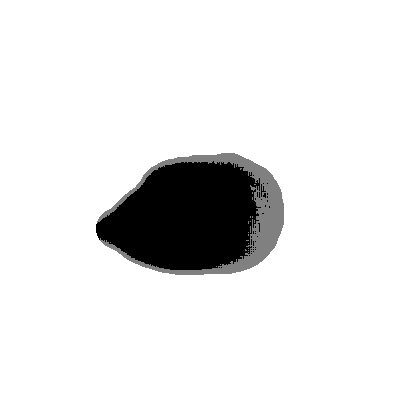}
			\caption{Spanning Trees.}
			\label{fig:ex3Spanning}
		\end{subfigure}
		\caption{Input image, ground truth and results from a variety algorithms for Example 3.}
		\label{fig:ex3}
	\end{figure}

\begin{figure}
		\centering
		\begin{subfigure}{0.32\textwidth}
			\includegraphics[width=\textwidth]{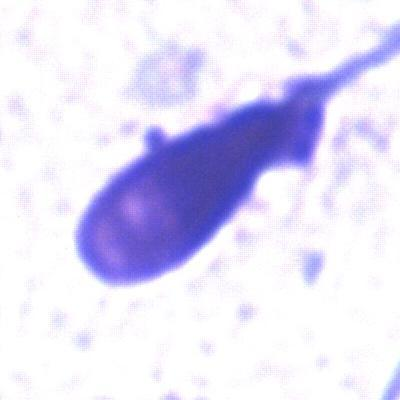}
			\caption{Input Image.}
			\label{fig:ex4in}
		\end{subfigure}
		\hfill
		\begin{subfigure}{0.32\textwidth}
			\includegraphics[width=\textwidth]{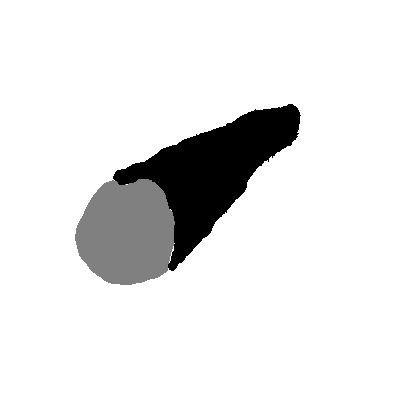}
			\caption{Ground Truth.}
			\label{fig:ex4GT}
		\end{subfigure}
		\hfill
		\begin{subfigure}{0.32\textwidth}
			\includegraphics[width=\textwidth]{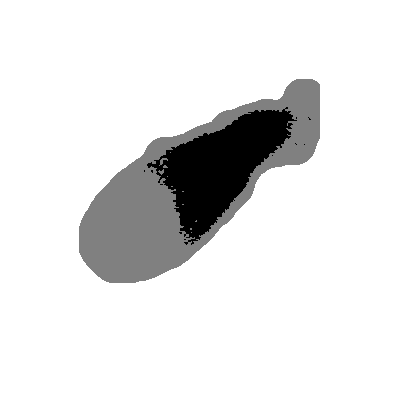}
			\caption{PDE Result.}
			\label{fig:ex4PDE}
		\end{subfigure}
		\hfill
		\begin{subfigure}{0.32\textwidth}
			\includegraphics[width=\textwidth]{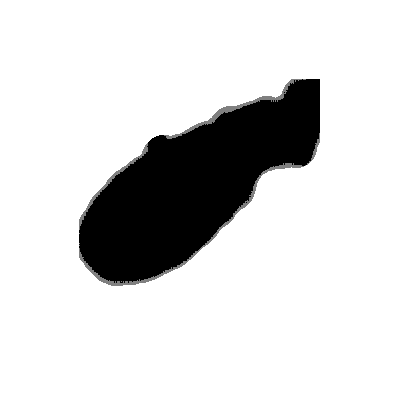}
			\caption{Agglomerative Clustering.}
			\label{fig:ex4Agg}
		\end{subfigure}
		\hfill
		\begin{subfigure}{0.32\textwidth}
			\includegraphics[width=\textwidth]{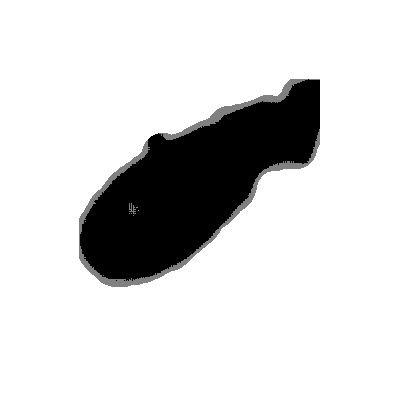}
			\caption{K Means Clustering.}
			\label{fig:ex4Kmean}
		\end{subfigure}
		\hfill
		\begin{subfigure}{0.32\textwidth}
			\includegraphics[width=\textwidth]{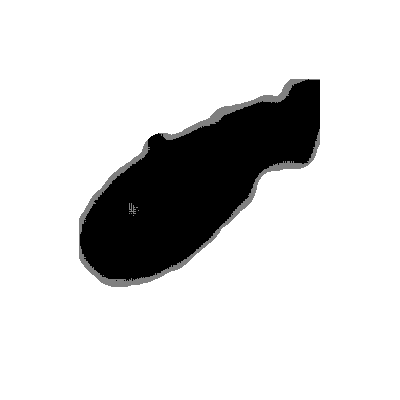}
			\caption{K Medoids Clustering.}
			\label{fig:ex4KMedoid}
		\end{subfigure}
		\hfill
		\begin{subfigure}{0.32\textwidth}
			\includegraphics[width=\textwidth]{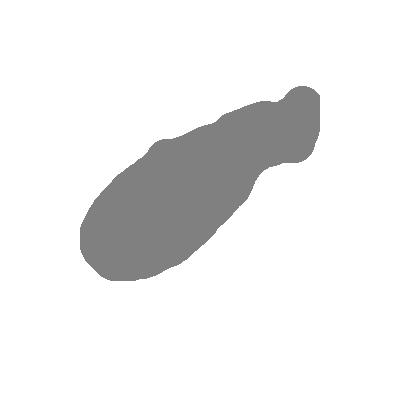}
			\caption{Spanning Trees.}
			\label{fig:ex4Spanning}
		\end{subfigure}
		\caption{Input image, ground truth and results from a variety algorithms for Example 4.}
		\label{fig:ex4}
	\end{figure}

\begin{figure}
		\centering
		\begin{subfigure}{0.32\textwidth}
			\includegraphics[width=\textwidth]{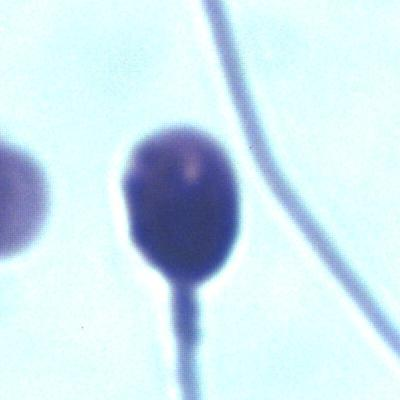}
			\caption{Input Image.}
			\label{fig:ex5in}
		\end{subfigure}
		\hfill
		\begin{subfigure}{0.32\textwidth}
			\includegraphics[width=\textwidth]{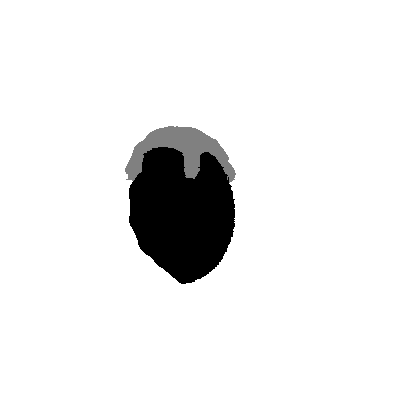}
			\caption{Ground Truth.}
			\label{fig:ex5GT}
		\end{subfigure}
		\hfill
		\begin{subfigure}{0.32\textwidth}
			\includegraphics[width=\textwidth]{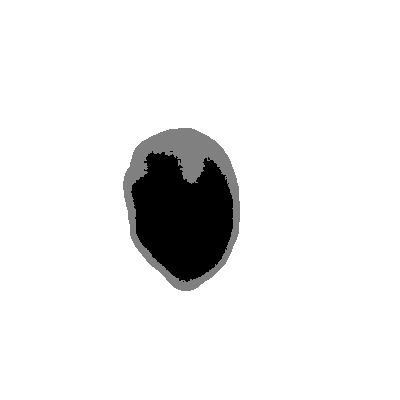}
			\caption{PDE Result.}
			\label{fig:ex5PDE}
		\end{subfigure}
		\hfill
		\begin{subfigure}{0.32\textwidth}
			\includegraphics[width=\textwidth]{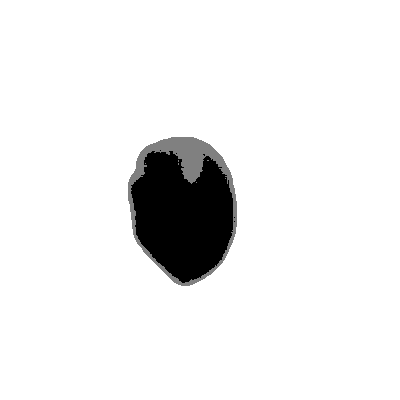}
			\caption{Agglomerative Clustering.}
			\label{fig:ex5Agg}
		\end{subfigure}
		\hfill
		\begin{subfigure}{0.32\textwidth}
			\includegraphics[width=\textwidth]{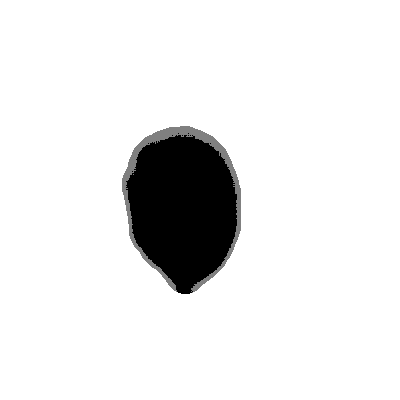}
			\caption{K Means Clustering.}
			\label{fig:ex5Kmean}
		\end{subfigure}
		\hfill
		\begin{subfigure}{0.32\textwidth}
			\includegraphics[width=\textwidth]{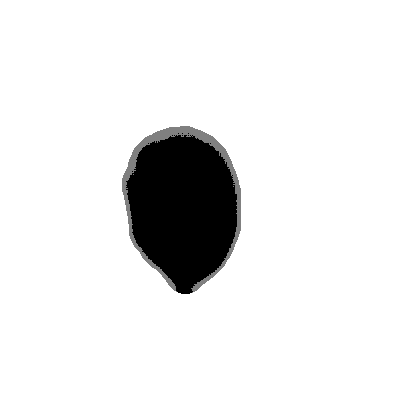}
			\caption{K Medoids Clustering.}
			\label{fig:ex5KMedoid}
		\end{subfigure}
		\hfill
		\begin{subfigure}{0.32\textwidth}
			\includegraphics[width=\textwidth]{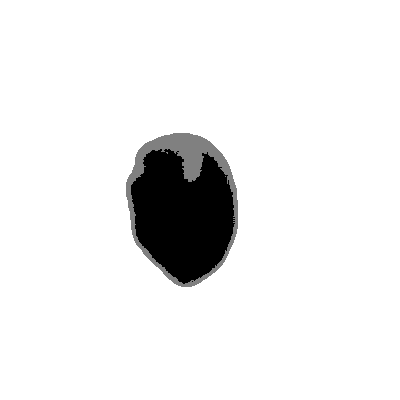}
			\caption{Spanning Trees.}
			\label{fig:ex5Spanning}
		\end{subfigure}
		\caption{Input image, ground truth and results from a variety algorithms for Example 5.}
		\label{fig:ex5}
	\end{figure}

	Figures \ref{fig:ex1}, \ref{fig:ex2}, \ref{fig:ex3}, \ref{fig:ex4} and \ref{fig:ex5} illustrate the classification outputs obtained using the present method as well as four standard unsupervised image segmentation algorithms; namely Agglomerative clustering, K-Means, K-Medoids and Spanning Trees. The average performance results for all examples of the present method and aforementioned methods are presented in Table \ref{tab:results}.

\begin{sidewaystable}
		\centering
		\caption{Table reporting the average $F_1$ scores and accuracy over all images for all methods.}\label{tab:results}
		\begin{tabular}{|ccccccc|}
			\hline
			\text{Method} & $F_1$\text{ Class 1} & $F_1$\text{ Class 2} & \text{Average }$F_1$ & \text{Accuracy Class 1} & \text{Accuracy Class 2} & \text{Average Accuracy} \\
			\hline
			\text{Agglomerate} & 0.214378 & 0.708983 & 0.46168 & 0.950434 & 0.946978 & 0.948706 \\
			\text{KMeans} & 0.45513 & 0.798757 & 0.626943 & 0.962562 & 0.966829 & 0.964696 \\
			\text{KMedoids} & 0.454328 & 0.79845 & 0.626389 & 0.962536 & 0.966779 & 0.964658 \\
			\text{PDE} & \textbf{0.784823} &\textbf{ 0.894653} & \textbf{0.839738} & \textbf{0.975313} & \textbf{0.98833} &\textbf{ 0.981821} \\
			\text{SpanningTree} & 0.45042 & 0.505314 & 0.289399 & 0.942096 & 0.963594 & 0.952845 \\
			\hline
		\end{tabular}
\end{sidewaystable}

\begin{sidewaystable}
	\centering
	\caption{Table reporting the accuracy for all images and all methods.}\label{tab:resultsAccuracy}
	\begin{tabular}{|ccccc|}
			\hline
			\text{Example} & \text{Method} & \text{Accuracy Class 1} & \text{Accuracy Class 2} & \text{Average Accuracy} \\
			\hline
			1 & \text{Agglomerate} & 0.92285 & 0.925719 & 0.924284 \\
			2 & \text{Agglomerate} & 0.956875 & 0.961444 & 0.959159 \\
			3 & \text{Agglomerate} & 0.947506 & 0.936206 & 0.941856 \\
			4 & \text{Agglomerate} & 0.93955 & 0.922138 & 0.930844 \\
			5 & \text{Agglomerate} & 0.985706 & 0.993394 & 0.98955 \\
			\hline
			1 & \text{KMeans} & \textbf{0.976012} & \textbf{0.984981} & \textbf{0.980497} \\
			2 & \text{KMeans} & 0.954438 & 0.964781 & 0.959609 \\
			3 & \text{KMeans} & 0.96795 & 0.973869 & 0.970909 \\
			4 & \text{KMeans} & 0.935137 & 0.927269 & 0.931203 \\
			5 & \text{KMeans} & 0.979019 & 0.986556 & 0.982788 \\
			\hline
			1 & \text{KMedoids} & \textbf{0.976012} & \textbf{0.984981} & \textbf{0.980497} \\
			2 & \text{KMedoids} & 0.954269 & 0.964781 & 0.959525 \\
			3 & \text{KMedoids} & 0.96795 & 0.973869 & 0.970909 \\
			4 & \text{KMedoids} & 0.935137 & 0.927269 & 0.931203 \\
			5 & \text{KMedoids} & 0.979056 & 0.986306 & 0.982681 \\
			\hline
			1 & \text{PDE} & 0.971231 & 0.97955 & 0.975391 \\
			2 & \text{PDE} &\textbf{ 0.981831} & \textbf{0.991081} &\textbf{ 0.986456} \\
			3 & \text{PDE} &\textbf{ 0.986413} & \textbf{0.996219} & \textbf{0.991316} \\
			4 & \text{PDE} & \textbf{0.95215} & \textbf{0.982944} & \textbf{0.967547} \\
			5 & \text{PDE} & 0.984938 & 0.991856 & 0.988397 \\
			\hline
			1 & \text{SpanningTree} & 0.924731 & 0.980788 & 0.952759 \\
			2 & \text{SpanningTree} & 0.935456 & 0.941144 & 0.9383 \\
			3 & \text{SpanningTree} & 0.962294 & 0.968025 & 0.965159 \\
			4 & \text{SpanningTree} & 0.901563 & 0.927925 & 0.914744 \\
			5 & \text{SpanningTree} & \textbf{0.986344} & \textbf{0.995337} & \textbf{0.990841} \\
			\hline
	\end{tabular}
\end{sidewaystable}

\begin{sidewaystable}
	\centering
	\caption{Table reporting the $F_1$ scores for all images and all methods.}\label{tab:resultsF1Score}
	\begin{tabular}{|ccccc|}
		\hline
		\text{Example} & \text{Method} & $F_1$\text{ Score Class 1} & $F_1$\text{ Score Class 2} & \text{Average }$F_1$\text{ Score} \\
		\hline
		1 & \text{Agglomerate} & 0.126645 & 0.611055 & 0.36885 \\
		2 & \text{Agglomerate} & 0.234184 & 0.750919 & 0.492551 \\
		3 & \text{Agglomerate} & 0.000713861 & 0.61277 & 0.306742 \\
		4 & \text{Agglomerate} & 0.0724971 & 0.648119 & 0.360308 \\
		5 & \text{Agglomerate} & 0.630473 & 0.948305 & 0.789389 \\
		\hline
		1 & \text{KMeans} & 0.835336 & \textbf{0.884941} & \textbf{0.860138} \\
		2 & \text{KMeans} & 0.308087 & 0.766715 & 0.537401 \\
		3 & \text{KMeans} & 0.597741 & 0.794272 & 0.696007 \\
		4 & \text{KMeans} & 0.123035 & 0.662412 & 0.392724 \\
		5 & \text{KMeans} & 0.397848 & 0.908775 & 0.653311 \\
		\hline
		1 & \text{KMedoids} & 0.835336 & \textbf{0.884941} & \textbf{0.860138} \\
		2 & \text{KMedoids} & 0.307561 & 0.766715 & 0.537138 \\
		3 & \text{KMedoids} & 0.597741 & 0.794272 & 0.696007 \\
		4 & \text{KMedoids} & 0.123035 & 0.662412 & 0.392724 \\
		5 & \text{KMedoids} & 0.394361 & 0.907244 & 0.650802 \\
		\hline
		1 & \text{PDE} & \textbf{0.839039} & 0.78767 & 0.813355 \\
		2 & \text{PDE} & \textbf{0.818005} & \textbf{0.92206} & \textbf{0.870033} \\
		3 & \text{PDE} & \textbf{0.87039}5 & \textbf{0.962067} & \textbf{0.916231} \\
		4 & \text{PDE} & \textbf{0.684471} & \textbf{0.865969} & \textbf{0.77522} \\
		5 & \text{PDE} & \textbf{0.712204} & 0.935498 & \textbf{0.823851} \\
		\hline
		1 & \text{SpanningTree} & 0 & 0.803125 & 0 \\
		2 & \text{SpanningTree} & 0.574828 & 0 & 0 \\
		3 & \text{SpanningTree} & 0.501858 & 0.759383 & 0.63062 \\
		4 & \text{SpanningTree} & 0.504093 & 0 & 0 \\
		5 & \text{SpanningTree} & 0.668688 & \textbf{0.964062} & 0.816375 \\
		\hline
	\end{tabular}
\end{sidewaystable}
	
\section{Conclusion}\label{sec:Conclusion}
	The present work extends the PDE based framework for image processing previously established for document image binarization \cite{jacobs2013novel,jacobs2015locally,jacobs2022unsupervised}. The application of CASA raises the issue of image trinarization, whereby an image data is classified into three distinct classes. This classification task is performed dynamically by the carefully constructed PDE model, which is shown to exhibit three steady states (corresponding to the three distinct classes). This proposed model is benchmarked against several other general purpose clustering algorithms and shown to perform exceptionally well for a variety of input images and is robust, even with fixed parameter values.\\
	\\
	The extensibility of the present image processing framework opens opportunities for novel processing algorithms, for new application domains. Specifically, the inclusion of local adaptivity \cite{jacobs2015locally}, or automatic control of model parameters for unsupervised methodologies \cite{jacobs2022unsupervised}, or edge-preserving image denoising \cite{guo2019nonlinear, perona1990scale}. A multitude of applications are available to this framework, and new architectures are necessary to be developed and explored to present novel solutions to arising problems.

\bibliographystyle{elsarticle-num} 
\bibliography{references.bib}

{\hyphenation{Post-Script Sprin-ger}}
\begin{thebibliography}{10}
\expandafter\ifx\csname url\endcsname\relax
  \def\url#1{\texttt{#1}}\fi
\expandafter\ifx\csname urlprefix\endcsname\relax\def\urlprefix{URL }\fi
\expandafter\ifx\csname href\endcsname\relax
  \def\href#1#2{#2} \def\path#1{#1}\fi

\bibitem{perona1990scale}
P.~Perona, J.~Malik, Scale-space and edge detection using anisotropic
  diffusion, IEEE Transactions on pattern analysis and machine intelligence
  12~(7) (1990) 629--639.

\bibitem{liu2011remote}
P.~Liu, F.~Huang, G.~Li, Z.~Liu, Remote-sensing image denoising using partial
  differential equations and auxiliary images as priors, IEEE geoscience and
  remote sensing letters 9~(3) (2011) 358--362.

\bibitem{kim2006pde}
S.~Kim, Pde-based image restoration: A hybrid model and color image denoising,
  IEEE Transactions on Image Processing 15~(5) (2006) 1163--1170.

\bibitem{catte1992image}
F.~Catt{\'e}, P.-L. Lions, J.-M. Morel, T.~Coll, Image selective smoothing and
  edge detection by nonlinear diffusion, SIAM Journal on Numerical analysis
  29~(1) (1992) 182--193.

\bibitem{jacobs2013novel}
B.~Jacobs, E.~Momoniat, A novel approach to text binarization via a
  diffusion-based model, Applied Mathematics and Computation 225 (2013)
  446--460.

\bibitem{jacobs2015locally}
B.~Jacobs, E.~Momoniat, A locally adaptive, diffusion based text binarization
  technique, Applied Mathematics and Computation 269 (2015) 464--472.

\bibitem{jacobs2018application}
B.~Jacobs, C.~Harley, Application of nonlinear time-fractional partial
  differential equations to image processing via hybrid laplace transform
  method, Journal of Mathematics 2018 (2018).

\bibitem{jacobs2022unsupervised}
B.~Jacobs, T.~Celik, Unsupervised document image binarization using a system of
  nonlinear partial differential equations, Applied Mathematics and Computation
  418 (2022) 126806.

\bibitem{weeratunga2003comparison}
S.~K. Weeratunga, C.~Kamath, Comparison of pde-based nonlinear anisotropic
  diffusion techniques for image denoising, in: Image Processing: Algorithms
  and Systems II, Vol. 5014, International Society for Optics and Photonics,
  2003, pp. 201--212.

\bibitem{chan2005image}
T.~F. Chan, J.~Shen, Image processing and analysis: variational, PDE, wavelet,
  and stochastic methods, SIAM, 2005.

\bibitem{hadri2021novel}
A.~Hadri, L.~Afraites, A.~Laghrib, M.~Nachaoui, A novel image denoising
  approach based on a non-convex constrained pde: application to ultrasound
  images, Signal, Image and Video Processing 15~(5) (2021) 1057--1064.

\bibitem{xing2011pde}
X.-X. Xing, Y.-L. Zhou, J.~S. Adelstein, X.-N. Zuo, Pde-based spatial
  smoothing: a practical demonstration of impacts on mri brain extraction,
  tissue segmentation and registration, Magnetic Resonance Imaging 29~(5)
  (2011) 731--738.

\bibitem{lu2009four}
B.~Lu, C.~Deng, Q.~Liu, J.~Li, Four order adaptive pde method for mri
  denoising, in: 2009 3rd International Conference on Bioinformatics and
  Biomedical Engineering, IEEE, 2009, pp. 1--4.

\bibitem{feng2019novel}
S.~Feng, A novel variational model for noise robust document image
  binarization, Neurocomputing 325 (2019) 288--302.

\bibitem{feng2022effective}
S.~Feng, Effective document image binarization via a convex variational level
  set model, Applied Mathematics and Computation 419 (2022) 126861.

\bibitem{wang2019indirect}
Y.~Wang, Q.~Yuan, C.~He, Indirect diffusion based level set evolution for image
  segmentation, Applied Mathematical Modelling 69 (2019) 714--722.

\bibitem{guo2020fourth}
J.~Guo, C.~He, Y.~Wang, Fourth order indirect diffusion coupled with shock
  filter and source for text binarization, Signal Processing 171 (2020) 107478.

\bibitem{zhang2020selective}
X.~Zhang, C.~He, J.~Guo, Selective diffusion involving reaction for
  binarization of bleed-through document images, Applied Mathematical Modelling
  81 (2020) 844--854.

\bibitem{nnolim2021dynamic}
U.~A. Nnolim, Dynamic selective edge-based integer/fractional-order partial
  differential equation for degraded document image binarization, International
  Journal of Image and Graphics (2021) 2250030.

\bibitem{nnolim2021improved}
U.~A. Nnolim, Improved integer/fractional order partial differential
  equation-based thresholding, Optik 229 (2021) 166265.

\bibitem{du2021nonlinear}
Z.~Du, C.~He, Nonlinear diffusion equation with selective source for
  binarization of degraded document images, Applied Mathematical Modelling 99
  (2021) 243--259.

\bibitem{macqueen1967some}
J.~MacQueen, et~al., Some methods for classification and analysis of
  multivariate observations, in: Proceedings of the fifth Berkeley symposium on
  mathematical statistics and probability, Vol.~1, Oakland, CA, USA, 1967, pp.
  281--297.

\bibitem{kaufman2009finding}
L.~Kaufman, P.~J. Rousseeuw, Finding groups in data: an introduction to cluster
  analysis, John Wiley \& Sons, 2009.

\bibitem{asano1988clustering}
T.~Asano, B.~Bhattacharya, M.~Keil, F.~Yao, Clustering algorithms based on
  minimum and maximum spanning trees, in: Proceedings of the fourth annual
  symposium on Computational Geometry, 1988, pp. 252--257.

\bibitem{rokach2005clustering}
L.~Rokach, O.~Maimon, Clustering methods, in: Data mining and knowledge
  discovery handbook, Springer, 2005, pp. 321--352.

\bibitem{kido2020quantification}
R.~Kido, Y.~Higo, Quantification of pore volume and degree of saturation in
  partially saturated sands with different bulk density, Japanese Geotechnical
  Society Special Publication 8~(6) (2020) 216--220.

\bibitem{kido2017evaluation}
R.~Kido, Y.~Higo, Evaluation of distribution of void ratio and degree of
  saturation in partially saturated triaxial sand specimen using micro x-ray
  tomography, Japanese Geotechnical Society Special Publication 5~(2) (2017)
  22--27.

\bibitem{otsu1975threshold}
N.~Otsu, A threshold selection method from gray-level histograms, Automatica
  11~(285-296) (1975) 23--27.

\bibitem{higo2014trinarization}
Y.~Higo, F.~Oka, R.~Morishita, Y.~Matsushima, T.~Yoshida, Trinarization of
  $\mu$x-ray ct images of partially saturated sand at different water-retention
  states using a region growing method, Nuclear Instruments and Methods in
  Physics Research Section B: Beam Interactions with Materials and Atoms 324
  (2014) 63--69.

\bibitem{mussel2016bitrina}
C.~M{\"u}ssel, F.~Schmid, T.~J. Bl{\"a}tte, M.~Hopfensitz, L.~Lausser, H.~A.
  Kestler, Bitrina—multiscale binarization and trinarization with quality
  analysis, Bioinformatics 32~(3) (2016) 465--468.

\bibitem{miahi2022genetic}
E.~Miahi, S.~A. Mirroshandel, A.~Nasr, Genetic neural architecture search for
  automatic assessment of human sperm images, Expert Systems with Applications
  188 (2022) 115937.

\bibitem{javadi2019novel}
S.~Javadi, S.~A. Mirroshandel, A novel deep learning method for automatic
  assessment of human sperm images, Computers in biology and medicine 109
  (2019) 182--194.

\bibitem{csavkay2014sperm}
O.~L. {\c{S}}avkay, E.~Cesur, M.~E. Yal{\c{c}}{\i}n, V.~Tav{\c{s}}ano{\u{g}}lu,
  Sperm morphology analysis with cnn based algorithms, in: 2014 14th
  International Workshop on Cellular Nanoscale Networks and their Applications
  (CNNA), IEEE, 2014, pp. 1--2.

\bibitem{prabaharan2021improved}
L.~Prabaharan, A.~Raghunathan, An improved convolutional neural network for
  abnormality detection and segmentation from human sperm images, Journal of
  Ambient Intelligence and Humanized Computing 12~(3) (2021) 3341--3352.

\bibitem{riegler2021artificial}
M.~A. Riegler, M.~H. Stensen, O.~Witczak, J.~M. Andersen, S.~Hicks, H.~L.
  Hammer, E.~Delbarre, P.~Halvorsen, A.~Yazidi, N.~Holst, et~al., Artificial
  intelligence in the fertility clinic: status, pitfalls and possibilities,
  Human Reproduction 36~(9) (2021) 2429--2442.

\bibitem{dai2021advances}
C.~Dai, Z.~Zhang, G.~Shan, L.-T. Chu, Z.~Huang, S.~Moskovtsev, C.~Librach,
  K.~Jarvi, Y.~Sun, Advances in sperm analysis: techniques, discoveries and
  applications, Nature Reviews Urology 18~(8) (2021) 447--467.

\bibitem{world2010laboratory}
W.~H. Organization, et~al., Who laboratory manual for the examination and
  processing of human semen (2010).

\bibitem{taha2015metrics}
A.~A. Taha, A.~Hanbury, Metrics for evaluating 3d medical image segmentation:
  analysis, selection, and tool, BMC medical imaging 15~(1) (2015) 1--28.

\bibitem{guo2019nonlinear}
J.~Guo, C.~He, X.~Zhang, Nonlinear edge-preserving diffusion with adaptive
  source for document images binarization, Applied Mathematics and Computation
  351 (2019) 8--22.

\end{thebibliography}

\end{document}